%% file: main.tex
\documentclass[letterpaper]{article}
\usepackage[preprint]{aaai2027}
\usepackage[hyphens]{url}
\usepackage{graphicx}
\usepackage{natbib}
\usepackage{caption}
\usepackage{amsmath}
\usepackage{amssymb}
\usepackage{booktabs}
\title{When Do Task Vectors Interfere? Mapping the Validity Boundaries of Weight-Space Composition}
\author{%
Chencheng Zhu\textsuperscript{\rm 1},
Xiaoyang Li\textsuperscript{\rm 2},
Taotao Cai\textsuperscript{\rm 2}
}
\affiliations{%
\textsuperscript{\rm 1}UNSW Sydney\\
\textsuperscript{\rm 2}University of Southern Queensland
}

\begin{document}
\maketitle

\begin{abstract}
Task arithmetic composes skills by adding weight displacements, and merged models are then judged on benchmark suites. We measure when that composition is functionally additive, and find that the answer depends as much on how the model is prompted as on which tasks are merged. Across two-dimensional composition surfaces---five model settings from 0.5B to 8B, two families, LoRA and full fine-tuning---pairwise non-additivity is real, seed-stable, and transfers in coarse order to unseen task pairs: all eight preregistered sign predictions held. But it is input-conditioned everywhere we measured: the same merged model that shows a six-point interaction contrast on code prompts shows none on math prompts, and wrapping the identical code prompts in the instruction template the adapters were trained on collapses the contrast twenty-fold, from $+6.9$ to $+0.3$ points---while re-serializing them in an untrained chat template leaves it intact ($+12.5$), falsifying our own preregistered prediction. Execution benchmarks (pass@1) inherit the training-format wrapper's blindness. Weight-space composition therefore supports coarse, input- and format-conditioned functional statements---not a universal merging-performance predictor, and not one that training-format evaluations can see.
\end{abstract}

\input{sections/01_introduction}
\input{sections/02_related_work}
\input{sections/03_method}
\input{sections/04_setup}
\input{sections/05_results}

\input{sections/06_discussion}
\input{sections/07_conclusion}

\bibliography{references}

\clearpage
\appendix
\input{sections/08_appendix}

\end{document}

%% file: sections/01_introduction.tex
\section{Introduction}

Task arithmetic offers an appealing interface for model reuse: subtract a shared base model from models fine-tuned for individual tasks, then add the resulting task vectors to combine capabilities~\cite{Ilharco2023}. In practice the merged model is then judged the way every model is judged---on benchmark suites, almost always through instruction-formatted prompts. This creates a blind spot that, to our knowledge, has not been measured: if how much interference a merge \emph{expresses} depends on how it is prompted, then a formatted evaluation can pass a merge whose interference is intact. This paper measures exactly that, and finds the blind spot is real---and precisely located: the instruction template the adapters were trained on collapses a twenty-fold interference contrast on identical prompts, while an untrained chat template leaves it intact.

The underlying issue is that a task vector is defined in parameter space, whereas success is judged by behavior, and the two geometries need not agree. Paths of low loss, symmetry alignment, and local linearization clarify weight connectivity and fine-tuning regimes~\cite{Garipov2018,Draxler2018,Entezari2022,Ainsworth2023,Malladi2023,OrtizJimenez2023}, but none implies that a task vector has a global meaning independent of the input, or that it composes benignly with another.

We therefore measure composition functionally. Scanning a two-dimensional coefficient surface over two task vectors, we compare the merged model's output distribution with an additive prediction assembled from the two axis paths, isolating the pairwise interaction after subtracting each axis's own nonlinearity (Figure~\ref{fig:method-overview}). The resulting interaction ratio is conditioned on a prompt distribution and estimated under matched seeds, norm interventions, and protocols fixed in advance.

Three findings emerge. \emph{First, functional interference is jointly determined by the task pair and the input}: after norm matching on Qwen2.5-1.5B, code+safety exceeds code+math by about six points on code prompts and more on instruction prompts, yet by nothing on math prompts. \emph{Second, the structure transfers coarsely}: in a six-task expansion whose comparison bins were fixed before the new adapters existed, all eight sign predictions on unseen task pairs held, and the primary contrast persists under full-parameter fine-tuning, scaling to 7B, and a Llama-3.1-8B cross-family audit. \emph{Third, expression is gated by training-format serialization}: raw public code, instruction, and safety prompts preserve the contrast, the training-format Alpaca wrapper erases it on the same code prompts, an untrained ChatML serialization does not, and execution-based pass@1 inherits the blindness. A companion causal study traces this gating to a denominator effect in the readout rather than a change in the underlying interaction~\cite{Zhu2026Gate}.

Concretely, we contribute:
\begin{itemize}
    \item \textbf{An input-conditioned interaction surface.} A functional estimand that separates the non-additivity \emph{between} two task vectors from the nonlinearity \emph{along} each, conditioned on the prompt distribution---turning ``do these vectors interfere?'' into a measurable, input-indexed quantity.
    \item \textbf{Coarse transferability.} Interference ordering predicted before the data existed: 8/8 sign predictions on unseen task pairs, with the primary contrast surviving full fine-tuning, two scales, and a second model family.
    \item \textbf{The format boundary, located.} Interference expression is gated by training-format serialization, not public provenance or chat formatting per se: a twenty-fold collapse on identical prompts under the training-format wrapper, no collapse under an untrained chat template ($+12.5$ points), and pass@1 inherits the blindness---so training-format evaluations cannot certify a merge interference-free.
\end{itemize}

%% file: sections/02_related_work.tex
\section{Related Work}

\paragraph{Task vectors and local linearity.}
Task vectors express fine-tuning as displacements that support addition and negation~\cite{Ilharco2023}. Weight averaging and mode-connectivity results characterize compatible basins and paths~\cite{Izmailov2018,Garipov2018,Draxler2018,Wortsman2022,Matena2022}, while permutation alignment extends compatibility across symmetric parameterizations~\cite{Entezari2022,Ainsworth2023}. Tangent-space arithmetic and kernel analyses connect composition to local function-space structure~\cite{OrtizJimenez2023,Malladi2023}; intrinsic-dimension and LoRA theory indicate that effective adaptation may occupy a small subspace~\cite{Aghajanyan2021,Jang2024LoRANTK}. Fine-tuned models are also known to occupy compact regions of weight space~\citep{Gueta2023Region}; our clustering audit therefore asks whether such regional structure predicts how two displacements interact, not whether it exists. We test whether such directions define stable behavioral axes across inputs.

\paragraph{Interference in model merging.}
TIES-Merging addresses redundant updates and sign disagreement~\cite{Yadav2023}; DARE sparsifies task deltas~\cite{Yu2024DARE}; AdaMerging learns coefficients~\cite{Yang2024}; and RegMean or MaTS use data statistics or task parameter subspaces~\cite{Jin2023RegMean,Tam2024MATS}. Other methods localize sparse weights~\cite{Wang2024TALL,He2024Localize}, modify feature alignment~\cite{Stoica2024ZipIt}, or guide task vector combination~\cite{Cheng2025WUDI}. Merging language models remains task dependent~\cite{Morrison2024}, and heterogeneous checkpoints violate assumptions of clean experts~\cite{Hitit2026InTheWild}. Two recent studies are closest to ours and differ in target. \citet{Zhou2026Mergeability} predict pairwise merge performance from interpretable checkpoint properties, and \citet{Sivaramakrishnan2026PermDoRA} show that simple parameter-space angles are not a stable predictor of adapter interference. We instead measure input-conditioned non-additivity in the space of output distributions; our external audits characterize where a benchmark-level performance claim would fail.

\paragraph{LoRA geometry and safety directions.}
Low-Rank Adaptation freezes base weights and trains low-rank factors~\cite{Hu2022}; its neural tangent optimization has been characterized theoretically~\cite{Jang2024LoRANTK}. AdapterFusion and AdapterSoup combine modular adaptations in activation or weight space~\cite{Pfeiffer2021AdapterFusion,Chronopoulou2023AdapterSoup}, and LoraHub assembles LoRA modules with learned coefficients for unseen tasks~\cite{Huang2024LoraHub}. For LoRA, KnOTS uses a shared SVD space~\cite{Stoica2025KnOTS}, while clustering by rank composes smaller semantic units~\cite{Zhao2025Lego}. Our audit with full parameter fine-tuning tests whether the ordering is an artifact of low-rank adaptation. Safe LoRA projects updates toward an aligned subspace~\cite{Hsu2024}.

\paragraph{Task relations.}
Task embeddings and grouping methods quantify which tasks may transfer or compete~\cite{Achille2019Task2Vec,Standley2020,Fifty2021}. These precedents motivate our prospectively specified family ordering, but our target differs: functional non-additivity after training rather than allocation during joint training.

\paragraph{Functional evaluation.}
Distances between output distributions provide continuous sensitivity even when task accuracy is discrete. For external behavioral evaluation, we use EvalPlus, which augments HumanEval and MBPP with additional tests to expose functionally incorrect code~\cite{Chen2021,Austin2021,Liu2023EvalPlus}. The discrepancy between our continuous interaction metric and pass@1 is itself one of the validity boundaries we characterize.

%% file: sections/03_method.tex
\section{Measuring Functional Interaction}

\paragraph{Task vector surface.}
Let $\theta_0$ be a shared base model and $\Delta_a=\theta_a-\theta_0$, $\Delta_b=\theta_b-\theta_0$ be LoRA displacements for two tasks after merging the adapters into the base weights. All quantities for a task are evaluated for a matched training seed $s$; we suppress this index until defining the estimand for each seed below. We evaluate
\begin{equation}
\theta(\alpha,\beta)=\theta_0+\alpha\Delta_a+\beta\Delta_b,
\quad \alpha,\beta\in[0,1.2]
\end{equation}
on a $7\times7$ grid. The point $(1,1)$ is the ordinary task arithmetic merge, while the axes retain the full nonlinear path for each task.

\begin{figure*}[t]
\centering
\includegraphics[width=\textwidth]{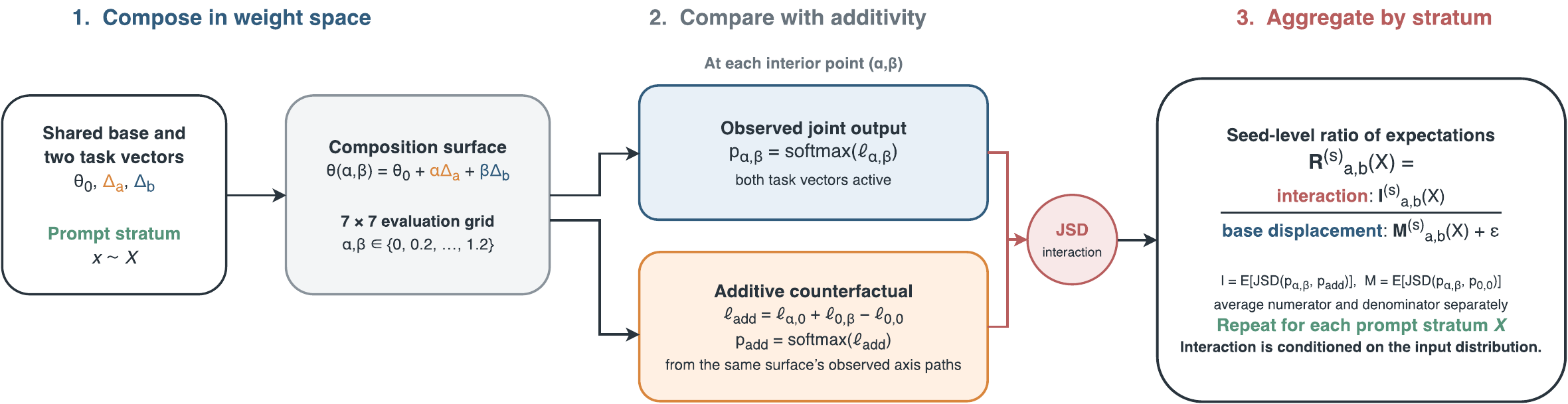}
\caption{Measurement pipeline for input-conditioned functional
interaction. Two task vectors from a shared base define a $7\times7$
composition surface that is evaluated on prompts $x\sim X$. At each interior
point, the observed joint distribution is compared with an additive
counterfactual constructed from the observed axis logits. Pointwise
Jensen--Shannon divergences measure joint interaction and displacement from
the base. The numerator and denominator are averaged separately over prompts
and the 36 interior points before forming a seed-level ratio of expectations.
Reported estimates then average the seed-level ratios with equal weight.}
\label{fig:method-overview}
\end{figure*}

For prompt $x$, denote the logits for the next token by $\ell_{\alpha,\beta}(x)$ and let
$p_{\alpha,\beta}(\cdot\mid x)=\operatorname{softmax}(\ell_{\alpha,\beta}(x))$.
A no-interaction prediction is assembled in logit space:
\begin{equation}
\ell_{\mathrm{add}}(x;\alpha,\beta)=
\ell_{\alpha,0}(x)+\ell_{0,\beta}(x)-\ell_{0,0}(x).
\end{equation}
Unlike a chord from the base to the merged endpoint, this prediction reuses both observed axis paths. This removes the marginal nonlinearity of each direction before measuring their joint effect. Softmax is applied only after logit composition, ensuring a normalized predictive distribution.

Equivalently, for vocabulary item $v$,
\begin{equation}
p_{\mathrm{add}}(v\mid x)\propto
\frac{p_{\alpha,0}(v\mid x)p_{0,\beta}(v\mid x)}
{p_{0,0}(v\mid x)}.
\end{equation}
This construction is a normalized additive displacement in log probability coordinates and is invariant to arbitrary scalar shifts of any constituent logit vector. It is a chosen operationalization of additivity between directions, not a definition of functional interaction that is independent of the model.

\paragraph{Interaction ratio conditioned on the input.}
The interaction for a prompt is
\begin{equation}
I_x(\alpha,\beta)=\operatorname{JSD}\!\left(
p_{\alpha,\beta}(\cdot\mid x),
p_{\mathrm{add}}(\cdot\mid x)
\right),
\end{equation}
where $\operatorname{JSD}$ is the Jensen--Shannon divergence, $p_{\mathrm{add}}=\operatorname{softmax}(\ell_{\mathrm{add}})$, and the
base displacement for a prompt is
$M_x(\alpha,\beta)=\operatorname{JSD}(p_{\alpha,\beta}(\cdot\mid x),
p_{0,0}(\cdot\mid x))$. Let
$\mathcal G_\circ=\{0.2,0.4,\ldots,1.2\}^2$ denote the 36 interior grid
points. To make the interaction comparable across surfaces with different
marginal movement, we normalize it by the actual displacement from the base:
\begin{equation}
\begin{aligned}
I^{(s)}_{a,b}(X)&=\mathbb{E}_{x\sim X,(\alpha,\beta)\sim\mathcal G_\circ}
I^{(s)}_x(\alpha,\beta),\\
M^{(s)}_{a,b}(X)&=\mathbb{E}_{x\sim X,(\alpha,\beta)\sim\mathcal G_\circ}
M^{(s)}_x(\alpha,\beta),\\
R^{(s)}_{a,b}(X)&=\frac{I^{(s)}_{a,b}(X)}
{M^{(s)}_{a,b}(X)+\varepsilon},\qquad \varepsilon=10^{-12},
\end{aligned}
\label{eq:ratio}
\end{equation}
where both expectations are uniform over prompts in $X$ and grid points in
$\mathcal G_\circ$, and all distributions in a given ratio come from matched
training seed $s$. The reported family estimate is
$\bar R_{a,b}(X)=|\mathcal S|^{-1}\sum_{s\in\mathcal S}R^{(s)}_{a,b}(X)$
with $\mathcal S=\{42,123,456\}$. The input distribution $X$ is an explicit
argument. We never aggregate it away before testing whether a pair
contrast changes across prompt strata.

\paragraph{Radial intervention.}
Update magnitude is a potential confound: a small vector may appear more additive merely because it remains closer to the base. Our primary mechanism analysis rescales every rank-16 task vector to the median norm of all vectors across tasks and seeds in the corresponding audit block. Thus, the target is computed separately for the calibration with four tasks, the expansion to six tasks, and each directed stress audit with three tasks. This \emph{core median norm matching} preserves directions while equalizing radii. These endpoints are interventions, not trained checkpoints; results at the natural radii remain secondary descriptions of training displacements.

\paragraph{Statistical estimands.}
Within each training seed, we separately average the interaction numerator and base displacement denominator over prompts and interior grid points, take their ratio, and then average the ratios across seeds with equal weight. We do not average ratios for individual prompts in the primary estimand and do not filter prompts by denominator size. Uncertainty is estimated with a hierarchical bootstrap that resamples seeds as the outer unit and prompts as the inner unit and recomputes the ratio of means. Matched contrasts share prompt indices across families. Because three seeds yield a minimum exact sign test value of $0.125$ under a directional alternative even with perfect concordance, we report bootstrap intervals over prompts and directional consistency across seeds separately. Nominal correlations across related task families are descriptive.

\paragraph{Normalization audit.}
Because $R$ can be unstable when the base displacement is small, a frozen reporting audit records denominator quantiles for every combination of family, stratum, and seed. It flags but never removes boundary cases. No primary cell crosses the specified boundary. The audit reproduces the ratio of expectations exactly and preserves its sign pattern when ratios are instead computed for each prompt before aggregation; the magnitudes depend on the estimator.

\paragraph{Diagnostics in parameter space.}
We audit global cosine similarity, the TIES sign conflict rate over coordinates that are nonnegligible in both vectors, and layerwise overlap between singular subspaces. These features test whether simple parameter summaries explain the observed functional surface; they are not used to define the primary interaction.

%% file: sections/04_setup.tex
\section{Experimental Setup}

\paragraph{Models and fine-tuning.}
We use Qwen2.5 base models at 0.5B, 1.5B, and 7B parameters~\cite{Qwen2025}. For each task and seed $\{42,123,456\}$, we train a rank-16 LoRA adapter on 3,000 examples for two epochs. LoRA is applied to the attention query, key, value, and output projections and the MLP gate, up, and down projections. We use $\alpha=32$, dropout $0.05$, learning rate $2\times10^{-4}$, cosine scheduling, $5\%$ warmup, effective batch size 16, and bfloat16. In a separate 0.5B audit, we fine-tune all parameters (Full FT) on math, code, and safety. This audit uses the same data, seeds, response-only objective, epochs, and effective batch size, with a prospectively specified learning rate of $2\times10^{-5}$. The audit on a second architecture repeats the directed math, code, and safety LoRA test on the Llama-3.1-8B base model with open weights~\cite{Dubey2024Llama}. We lock the checkpoint revision before training, and the analysis computes LoRA inner products directly from adapter factors and composes two adapters during the forward pass.

The frozen protocol supervises response tokens and the true EOS token while masking prompt and padding positions. Only checkpoints trained under this response-only protocol enter the main study. An earlier pilot protocol shared the padding and EOS identifiers, which inadvertently masked real EOS targets; it was replaced before any surface reported here was evaluated, and no pilot checkpoint enters any reported estimate.

\paragraph{Tasks and prompts.}
The four calibration tasks are math (GSM8K;~\citealt{Cobbe2021}), code (CodeAlpaca-20k;~\citealt{Chaudhary2023}), general instruction following (Alpaca;~\citealt{Taori2023}), and safety response modeling (PKU-SafeRLHF;~\citealt{Ji2024}). The prospectively specified expansion adds XSum summarization~\cite{Narayan2018} and OPUS Books translation from English to French~\cite{Tiedemann2012}. Long XSum documents are truncated inside the document field while preserving the response delimiter. Each controlled stratum contains 60 prompts held out from training and is evaluated independently.

\paragraph{Calibration and confirmatory roles.}
The initial 1.5B block with four tasks is exploratory calibration. It evaluates three anchor families: code+math, code+safety, and instruction+safety. The central code+safety versus code+math comparison is useful because both families share the same code checkpoint within each seed; its 1.5B result is descriptive rather than prospective. We froze this comparison before the 0.5B replication and subsequent audits of parameterization, scale, and model family. The expansion to six tasks evaluates nine out-of-sample (OOS) families involving a task held out from calibration plus the three anchors, for 12 families and 36 surfaces across families and seeds. Table~\ref{tab:blocks} in the appendix summarizes the experimental blocks; it counts selected surfaces for each family and seed, not every possible task pair.

\paragraph{Prospective specifications.}
Before training summarization and translation adapters, we froze coarse interaction bins. Pairing either unseen task with safety or instruction was predicted high; pairing with code was predicted middle; pairing with math was predicted low. Summarization+translation was registered as a bridge above the low bin. Success required at least $75\%$ of comparisons between unseen high and low pairs to have the predicted sign. Before training 0.5B adapters, we separately registered that code+safety would exceed code+math on code and instruction prompts, with at least two of three seed differences positive per stratum. Before the Full FT, Qwen2.5-7B, and Llama-3.1-8B audits, we froze the same two primary strata and required all three seed differences to be positive with the interval from the hierarchical bootstrap excluding zero for a strong persistence result. The 7B and Llama tests evaluated two LoRA adapters directly after confirming equivalence between dense and adapter implementations on the 1.5B pipeline.

The OOS bins encode a directional hypothesis from calibration: the new summarization and translation tasks should interact more with instruction or safety than with math, with code intermediate. The dated local specification and its complete trace from prediction to result are included in the ancillary files. Because no immutable external registry is available, we describe the test as prospectively specified rather than externally registered.

\paragraph{External evaluation.}
We evaluate transfer to public prompts at three levels. First, the raw public strata use EvalPlus HumanEval+ and MBPP+ prompt fields for code~\cite{Liu2023EvalPlus}, the GSM8K test split for math~\cite{Cobbe2021}, AlpacaEval instructions~\cite{Dubois2024AlpacaEval}, and a stratified XSTest sample with 30 safe and 30 unsafe prompts for safety~\cite{Rottger2024XSTest}. Each stratum contains 60 prompts sampled with a fixed seed and recorded with dataset revisions and prompt hashes. Second, a format bridge diagnostic reruns the identical 60 public code prompts inside the Alpaca instruction wrapper used by response-only training; a follow-up reruns them in the model's native ChatML chat template, which the adapters were never trained on, with the same instruction sentence. This follow-up, the ratio anchors, and the TIES operator probe reported in Results share a single preregistration, written after the primary studies froze; each records a directional prediction and its outcome. Third, for execution behavior, we generate greedy code on all 164 HumanEval+ and 378 MBPP+ prompts, save raw generations and audits of EOS and length caps, sanitize with the official EvalPlus pipeline, and evaluate pass@1. Continuous and discrete evaluations answer different questions and are not substituted for one another.

%% file: sections/05_results.tex
\section{Results}

The three findings of the introduction organize this section---where non-additivity appears, whether its ordering transfers to unseen pairs, and what evaluations can see it---followed by robustness and ablation audits that exclude alternative explanations. Throughout, point estimates are contrasts $C=R_{\text{code+safety}}-R_{\text{code+math}}$ in percentage points with 95\% hierarchical-bootstrap intervals; the statistical status of each analysis (inferential, descriptive, or null) is stated once where it is reported, and raw estimator components appear in the ancillary files.

\subsection{Functional Interference Depends on the Input}

\begin{figure*}[t]
\centering
\includegraphics[width=\textwidth]{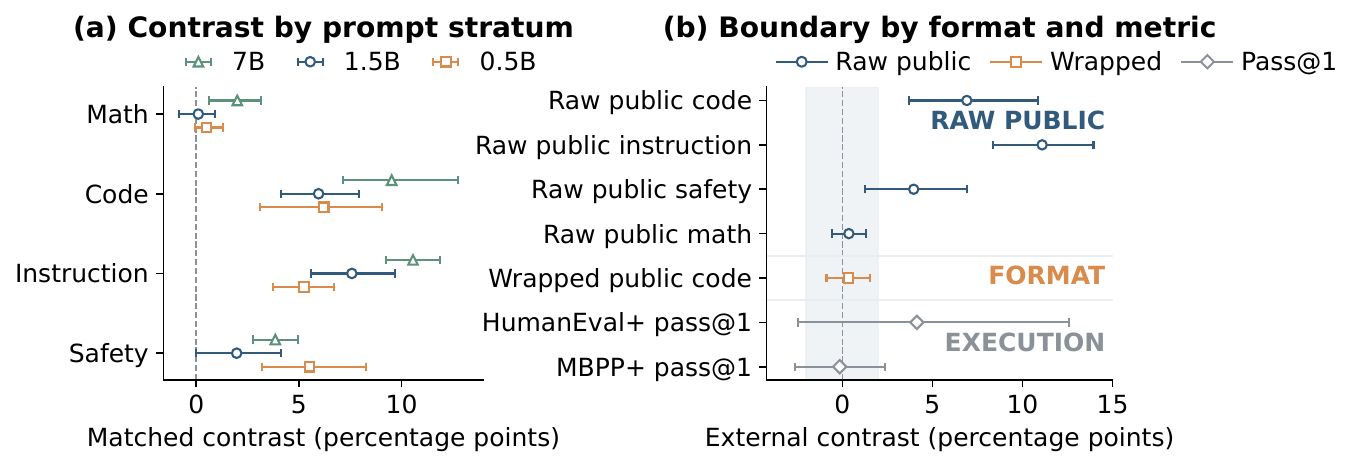}
\caption{Functional interference depends on input stratum, prompt format, and evaluation metric.
(a) Contrasts between code+safety and code+math after norm matching across controlled prompt strata for Qwen2.5 models at 0.5B, 1.5B, and 7B. Points are means and bars are 95\% intervals from a hierarchical bootstrap over prompts within seeds ($n=3$ training seeds).
(b) Raw public prompts preserve the continuous contrast on code, instruction, and safety inputs, whereas an Alpaca wrapper collapses the contrast on the same public code prompts. Pass@1 does not robustly reproduce the contrast. The shaded band is the $[-2,+2]$-point equivalence region of Table~\ref{tab:external}.}
\label{fig:boundaries}
\end{figure*}

Norm matching does not remove the family contrast, but the input does. On Qwen2.5-1.5B (Table~\ref{tab:main}; Figure~\ref{fig:boundaries}a), code+safety exceeds the matched code+math control by $+6.0$ points on code prompts and $+7.6$ on instruction prompts---and by $+0.1$ on math prompts, a null. Neither pair identity nor input distribution alone determines compositionality.

\begin{table*}[t]
\centering
\small
\setlength{\tabcolsep}{4.5pt}
\begin{tabular}{lrrrrr}
\toprule
 & \multicolumn{3}{c}{Qwen2.5 LoRA} & \multicolumn{1}{c}{Qwen2.5 Full FT} & \multicolumn{1}{c}{Llama-3.1 LoRA} \\
\cmidrule(lr){2-4}\cmidrule(lr){5-5}\cmidrule(lr){6-6}
Stratum & 0.5B & 1.5B & 7B & 0.5B & 8B \\
\midrule
Math & $+0.50$ $[-0.08,1.31]$ & $+0.10$ $[-0.83,0.91]$ & $+2.00$ $[0.62,3.16]$ & --- & --- \\
Code & $+6.21$ $[3.11,9.04]$ & $+5.96$ $[4.14,7.93]$ & $+9.52$ $[7.14,12.77]$ & $+4.37$ $[2.58,6.27]$ & $+4.87$ $[3.86,5.88]$ \\
Instruction & $+5.26$ $[3.74,6.70]$ & $+7.58$ $[5.59,9.69]$ & $+10.56$ $[9.23,11.88]$ & $+2.98$ $[2.43,3.54]$ & $+5.30$ $[4.03,6.59]$ \\
Safety & $+5.52$ $[3.21,8.26]$ & $+1.97$ $[-0.03,4.14]$ & $+3.85$ $[2.78,4.96]$ & --- & --- \\
\bottomrule
\end{tabular}
\caption{Main result: the matched contrast $C=R_{\text{code+safety}}-R_{\text{code+math}}$ (Eq.~\ref{eq:ratio}) in percentage points, across prompt strata and model settings, with 95\% hierarchical-bootstrap intervals. Qwen2.5 LoRA columns share the rank-16 protocol; the Full FT and Llama audits registered only the code and instruction strata (dashes: outside protocol). Component ratios for the 1.5B calibration appear in Table~\ref{tab:e0-decomposition}.}
\label{tab:main}
\end{table*}

\subsection{Prospective Directional Test on Unseen Pairs}

The six-task expansion trains 18 checkpoints and evaluates 12 task-pair families across three seeds (36 surfaces); nine families contain a task held out from calibration. The comparison bins and success rule were frozen in advance (see Experimental Setup), with a machine-readable trace from prediction to result in the ancillary files.

The predictions held: \textbf{all 8/8 high-versus-low comparisons on unseen pairs have the predicted sign} when averaged across six strata (Table~\ref{tab:oos}), 12/12 including the frozen low anchor, against a preregistered 75\% threshold. Because pair families share checkpoints, we report this as prospective directional evidence rather than as a definitive test. The supporting randomization permutes only the task labels that define the bins, holding every trained checkpoint fixed; it admits just twelve distinct assignments, of which the observed one is the unique 8/8, so the exact $p=0.083$ is the smallest value the design can produce---finite-sample resolution, not weak agreement. Finer structure is weaker: high exceeds middle in 5/8, and the ordinal correlation over held-out families is $\rho=0.57$ ($n=9$; $0.70$ with calibration anchors, a secondary analysis since the anchors define the bins).

Transfer remains input-conditioned: the high-versus-low sign accuracy is 100\% on instruction, safety, and summarization prompts, 75\% on translation, but 50\% on code and 37.5\% on math. The aggregate ordering is not a global relation in parameter space.

The stratum columns of Table~\ref{tab:oos} connect the first two findings: the ordering transfers exactly on the inputs where interference is expressed in a family-general way. The math column is flat---merged models remain near-additive there regardless of pair, so there is no ordering to detect; code expresses strong anchor interference ($+6.0$ points) but pair-idiosyncratically, and the bins do not rank unseen pairs there. Instruction, safety, and summarization are the regime where interference is both expressed and family-general, and there the ordering transfers perfectly. Input-conditioning is therefore not a nuisance dimension to average away; it determines where prediction is possible at all.

\begin{table*}[t]
\centering
\footnotesize
\setlength{\tabcolsep}{4.5pt}
\begin{tabular}{llrrrrrrr}
\toprule
Family & Bin & Code & Instr. & Math & Safety & Summ. & Trans. & Mean \\
\midrule
safety+translation & High & 4.88 & 11.45 & 3.79 & 10.03 & 38.71 & 30.85 & 16.62 \\
summarization+translation & Bridge & 9.42 & 15.07 & 4.99 & 8.42 & 29.73 & 5.67 & 12.22 \\
code+summarization & Middle & 3.00 & 10.51 & 3.31 & 12.33 & 22.73 & 14.70 & 11.10 \\
instruction+summarization & High & 3.43 & 12.09 & 3.71 & 12.40 & 22.01 & 12.26 & 10.98 \\
instruction+translation & High & 3.17 & 9.14 & 3.30 & 12.25 & 22.34 & 11.90 & 10.35 \\
safety+summarization & High & 4.38 & 11.85 & 3.66 & 8.67 & 22.81 & 5.04 & 9.40 \\
math+summarization & Low & 6.29 & 5.41 & 3.59 & 5.34 & 20.33 & 10.14 & 8.52 \\
code+translation & Middle & 2.58 & 6.92 & 2.94 & 10.18 & 13.46 & 6.38 & 7.08 \\
math+translation & Low & 2.89 & 6.04 & 3.90 & 4.41 & 14.07 & 5.04 & 6.06 \\
\bottomrule
\end{tabular}
\caption{The nine unseen families on the 1.5B surface after norm matching: interaction ratio $R$ (\%) for each prompt stratum, averaged over three seeds, with the six-stratum mean that the registered bins predict. Rows are sorted by the mean; bins were registered before the summarization and translation adapters existed. The stratum columns show the input-conditioning directly: the mean ordering is carried by the instruction, safety, and summarization columns, while math is flat and code is pair-idiosyncratic.}
\label{tab:oos}
\end{table*}

\begin{table*}[t]
\centering
\small
\setlength{\tabcolsep}{5pt}
\begin{tabular}{lrrr}
\toprule
Condition & Raw $C$ & Norm $C$ [95\% CI] & Outcome \\
\midrule
raw math & $+0.01$ & $+0.36$ $[-0.59,1.31]$ & strong eq. \\
raw code & $+8.09$ & $+6.91$ $[3.71,10.87]$ & residual \\
raw instruction & $+10.17$ & $+11.09$ $[8.34,13.94]$ & residual \\
raw safety & $+3.16$ & $+3.96$ $[1.24,6.93]$ & residual \\
wrapped code (Alpaca) & $+1.00$ & $+0.34$ $[-0.16,0.86]$ & strong eq. \\
wrapped code (ChatML) & $+11.76$ & $+12.51$ $[7.70,20.87]$ & residual \\
pass@1 (HE+) & n/a & $+4.14$ $[-2.44,12.60]$ & CI includes 0 \\
pass@1 (MBPP+) & n/a & $-0.15$ $[-2.65,2.38]$ & CI includes 0 \\
\bottomrule
\end{tabular}
\caption{The format boundary in percentage points. Raw $C$ is the matched contrast at natural checkpoint radii; Norm $C$ applies core median norm matching. Continuous intervals use a hierarchical bootstrap over prompts within seeds; pass@1 intervals use a hierarchical bootstrap over tasks within seeds. ``Strong eq.'' requires the hierarchical 90\% interval and every per-seed 90\% interval to lie inside the equivalence region $[-2,+2]$. The ChatML row is a separately preregistered follow-up.}
\label{tab:external}
\end{table*}

\subsection{The Format Boundary: Wrappers Hide Interference}

The paper's sharpest result is what happens on \emph{identical} prompts under different serialization (Table~\ref{tab:external}; Figure~\ref{fig:boundaries}b). Raw public code, instruction, and safety prompts preserve the contrast ($+6.9$, $+11.1$, $+4.0$ points after norm matching), and public math behaves as the intended low anchor ($+0.4$, satisfying the equivalence gate)---so public provenance does not attenuate the signal. But when the same 60 public code prompts are wrapped in the Alpaca instruction template---the serialization the adapters were trained on---the contrast collapses twenty-fold, from $+6.9$ to $+0.3$ points, and satisfies the strong equivalence gate. Re-serializing the identical prompts in the model's native ChatML chat template, which these adapters were never trained on, does the opposite: the contrast survives at $+12.5$ points---larger than raw---falsifying our preregistered prediction that it too would collapse. \textbf{The boundary is the training format: templates the adapters were trained on hide the interference; an untrained serialization does not.} A companion causal study supplies the mechanism and brackets the boundary from both sides: two further training-format phrasings inflate the main-effect denominator $11$--$35\times$ and suppress the expressed ratio in 18/18 cells, while a length-matched non-instruction prefix moves numerator and denominator together and hides nothing---the class ChatML matches~\cite{Zhu2026Gate}.

Two further probes bound what evaluations can see. A teacher-forcing check over reference continuations finds the code contrast intact over sequences ($+3.7$) and a small residual on math ($+1.9$; intervals in the appendix), so the math null is specific to the first-token estimand. Execution behavior is weaker still: the pass@1 interaction contrasts on HumanEval+ ($+4.1$) and MBPP+ ($-0.1$) are both compatible with zero and with moderate effects (Table~\ref{tab:external}). Together with the wrapper collapse, this is the deployment-relevant boundary: \emph{training-format and execution-based evaluations, as commonly run, do not see what the raw-prompt surface measures}.

\subsection{Robustness and Ablation Audits}

Estimator-level controls---vector norm, denominator behavior, and a cached recomputation of the primary estimator---are deferred to the appendix; none changes any conclusion. The two audits below ask whether the effect is a training artifact and whether parameter summaries could have predicted it.

\paragraph{Parameterization, scale, and family.}
The contrast survives every audit that targets an alternative explanation: on the registered code and instruction strata, every audit shows 3/3 seed differences positive with intervals excluding zero (Table~\ref{tab:main}). Full-parameter fine-tuning at 0.5B removes the rank-16 LoRA parameterization: the ordering persists at reduced magnitude, so parameterization modulates the size of the effect but does not create it. Scaling to Qwen2.5-7B \emph{increases} both contrasts---and at 7B even the math stratum turns weakly positive ($+2.0$)---which is inconsistent with a simple capacity-bottleneck account. Llama-3.1-8B transfers the effect to a second model family under the strong rule. The scale trend is itself stratum-specific: while code and instruction contrasts grow with model size, the safety-stratum contrast is larger at 0.5B ($+5.5$ points) than at 1.5B ($+2.0$, with an interval crossing zero; Table~\ref{tab:main}). A uniform capacity story would not produce opposite orderings on different inputs; this is the input-conditioning of the main finding expressed along the scale axis.

\paragraph{Anchoring the ratio.}
Two anchors give the ratio a scale (descriptive means; 1.5B, rank-16, code vectors). A random direction of matched norm composes almost perfectly additively with every code vector ($R=0.1$--$0.5\%$ on all four strata). Two seeds of the \emph{same} task compose least additively of anything we measured ($9.3\%$ mean on code prompts, exceeding code+safety)---falsifying our prediction that same-task pairs would sit near the floor. Because the composed point doubles the displacement along one direction, the same-task anchor mixes relatedness with curvature; we read it as an upper anchor only. On this scale the cross-task contrasts are interior points, and $R$ rises monotonically with functional relatedness: unrelated $0.4\%$, code+math $2.5\%$, code+safety $8.5\%$, same-task $9.3\%$.

\paragraph{Operator scope.}
A TIES follow-up locates a boundary of the estimand itself. Composing the same vectors by trim, sign election, and disjoint mean shrinks the joint displacement to 40--70\% of its additive counterpart, and the additive counterfactual then overshoots everywhere: the discrepancy term inflates $1.4$--$7\times$, largest where main effects are smallest, and its contrasts stop tracking the input pattern (all three registered checks failed). The additive counterfactual is a valid reference only for additive composition; TIES's first-order effects still show the input pattern descriptively, and interference under a non-additive operator needs an operator-consistent counterfactual, left as future work.

\paragraph{Parameter-space summaries.}
Parameter geometry tells a different story. At 1.5B the 18 task directions show weak but detectable clustering (all three diagnostics significant; Table~\ref{tab:param-audit}); at 0.5B the same audit detects none---yet the functional contrast there is intact. \textbf{Functional structure persists where this parameter-clustering audit detects nothing}, which is the concrete sense in which parameter summaries under-determine functional composition. Consistently, static parameter and data-similarity summaries failed to predict held-out families (negative results in the ancillary files).

%% file: sections/06_discussion.tex
\section{Discussion and Limitations}

\paragraph{What the results establish.}
Task-vector interference is a measurable, seed-stable functional quantity with three properties the field's parameter-space picture does not predict: it is jointly determined by the task pair and the input distribution; its coarse ordering is predictable for unseen pairs, before their adapters exist, across parameterizations, scales, and one additional family; and its \emph{expression} is gated specifically by training-format serialization, so the evaluations most commonly run on merged models---harnesses in the tuning template and execution benchmarks---can report a clean merge whose interference is intact on raw inputs. A companion causal study takes the next step, tracing the wrapper collapse to a denominator effect in the readout and the interference itself to the marginal displacements rather than their interaction~\cite{Zhu2026Gate}.

\paragraph{Evaluating a merge for interference.}
The findings translate into a concrete protocol. (1) Probe the merge on raw, unwrapped prompts from the input families the deployment will face: a harness in the tuning template can certify a merge that carries intact interference on raw inputs. (2) Measure against a matched control---a second pair sharing a checkpoint, evaluated at matched vector norms---and report the contrast between pairs rather than an absolute interaction value. (3) Keep behavioral benchmarks as deployment gates, not interference measurements: pass@1 was compatible with both zero and moderate interference.

\paragraph{Scope and statistical basis.}
The unseen-pair score is directional evidence, not a definitive inference (see Results). Three seeds check directional consistency only. Controlled prompts are diagnostic instruments, and the first-token estimand does not by itself predict decoded behavior. Norm-matched endpoints are interventions. The ChatML and anchor follow-ups cover 1.5B only (code stratum; code vectors). Full fine-tuning is audited only at 0.5B; Llama-3.1-8B covers three tasks. Task labels describe training data, not verified capability. The 0.5B clustering audit bounds what \emph{this} audit detects, not what parameter structure exists.

\paragraph{Broader impact.}
Unreliable composition can alter capabilities or safety behavior. Our measurements are diagnostics, not safety guarantees: including a safety-trained adapter does not establish performance on harmful compliance or excessive refusal, and deployment requires behavioral and safety evaluation for the intended application.

\paragraph{Artifacts.}
The ancillary files contain the dated protocols, prompt hashes, machine-readable audits, and model revision pins; Llama weights are not redistributed.

%% file: sections/07_conclusion.tex
\section{Conclusion}

\looseness=-1
Task-vector interference is real, measurable, and coarsely predictable: sign predictions on unseen pairs held 8/8 and the primary contrast survived every stress audit. Yet the same merge measures clean or conflicted depending on the input, and the training-format wrapper hides a twenty-fold contrast that an untrained chat template leaves intact---a blindness execution benchmarks inherit. Task-vector composition is an input-conditioned functional geometry with measured validity boundaries---here, for additive composition---not a universal semantic coordinate system; evaluations that never leave the wrapper never see the difference.

%% file: sections/08_appendix.tex
\section{Expanded Methods and Reproducibility}

\subsection{Additive Counterfactual in Logit Space}

For a prompt $x$, let $\ell_{\alpha,\beta}(x)$ be the logits at the first token for
$\theta_0+\alpha\Delta_a+\beta\Delta_b$.  The no-interaction counterfactual is

\begin{equation}
\ell_{\mathrm{add}}(x;\alpha,\beta)
=\ell_{\alpha,0}(x)+\ell_{0,\beta}(x)-\ell_{0,0}(x).
\end{equation}

Softmax is applied only after this construction in logit space, ensuring that the
counterfactual is a normalized distribution.  The interaction for a prompt is
$I_x(\alpha,\beta)=\operatorname{JSD}(p_{\alpha,\beta},p_{\mathrm{add}})$.
The denominator of the reported ratio is
$M_x(\alpha,\beta)=\operatorname{JSD}(p_{\alpha,\beta},p_{0,0})$.
We average $I_x$ and $M_x$ separately over prompts and interior grid points,
then report their ratio; we do not average ratios for individual prompts with unstable
small denominators.

The grid is exactly
$\{0.0,0.2,0.4,0.6,0.8,1.0,1.2\}$ on each axis. We cache the base and
logits along each axis and evaluate all points on the two dimensional surface, but only points with
$\alpha>0$ and $\beta>0$ enter the primary surface mean. Interior grid points
and prompts receive equal weight within each seed. JSD at the first token uses the
complete vocabulary and natural logarithms, so it is reported in nats; no
filtering based on the denominator is applied. The core median intervention rescales
each displacement using its global Frobenius norm, computed by summing squared
entries over all included parameter tensors, to the median norm of the core set
of rank-16 checkpoints before evaluating the surface. Deltas from fine-tuning
of all parameters include embeddings, normalization parameters, and the tied language
model head exactly once after deduplicating shared parameters.

\begin{table}[t]
\centering
\footnotesize
\begin{tabular}{lrrr}
\toprule
Study & Tasks & Seeds & Surfaces \\
\midrule
1.5B calibration & 4 & 3 & 9 \\
1.5B OOS & 6 & 3 & 36 \\
0.5B replication & 4 & 3 & 9 \\
0.5B Full FT audit & 3 & 3 & 6 \\
Qwen2.5-7B scale test & 3 & 3 & 6 \\
Llama-3.1-8B architecture test & 3 & 3 & 6 \\
Public raw/wrapped prompts & 2 fam. & 3 & 6 \\
\bottomrule
\end{tabular}
\caption{Main experimental blocks. Each surface is evaluated separately on every available prompt stratum.}
\label{tab:blocks}
\end{table}

\subsection{Training Protocol for Response Tokens}

Table~\ref{tab:full-config} lists the complete training and evaluation configuration. An earlier pilot protocol shared the padding and EOS identifier, which
inadvertently masked real EOS targets. The corrected final protocol,
\texttt{response-only-eos-v2-}\allowbreak\texttt{20260712}, tokenizes the prompt
and response separately, masks every prompt and padding position in the label
sequence, and supervises the response together with a true EOS token.  Each
encoded example is checked for four invariants: prompt labels are masked,
response labels are nonempty, padding labels are masked, and the final real
input and label tokens equal EOS.  If a response is truncated, its last retained
token is replaced with EOS.  Long XSum documents are truncated inside the source
field before the Alpaca template is assembled, preserving the response marker.

\begin{table}[t]
\centering
\footnotesize
\begin{tabular}{ll}
\toprule
Item & Frozen value \\
\midrule
Base models & Qwen2.5-0.5B, Qwen2.5-1.5B \\
Examples per task & 3,000 \\
Epochs & 2 \\
Optimizer schedule & cosine; 5\% warmup \\
Learning rate & $2\times10^{-4}$ \\
Device / effective batch & 4 / 16 \\
LoRA rank / $\alpha$ / dropout & 16 / 32 / 0.05 \\
Target modules & q,k,v,o,gate,up,down projections \\
Seeds & 42, 123, 456 \\
Composition grid & $7\times7$, scales 0 to 1.2 \\
Controlled prompts & 60 per stratum \\
\bottomrule
\end{tabular}
\caption{Frozen configuration for the calibration and replication blocks. The Qwen2.5-7B and Llama-3.1-8B audits reuse this recipe on their own base models under separately frozen protocols.}
\label{tab:full-config}
\end{table}

\subsection{Statistical Estimation}

Uncertainty intervals use a hierarchical bootstrap with the training seed as
the outer unit and prompts sampled within seed. Matched family contrasts reuse
the same prompt indices because both families are evaluated on the same prompt
set. Reported 95\% intervals are the 2.5th and 97.5th percentiles of the
bootstrap replicates. We additionally report the three differences across seeds.
With only three seeds, an exact sign test cannot yield $p<0.05$ under a
directional alternative, even when all signs agree. The bootstrap intervals
quantify uncertainty across the prompt distribution and do not turn prompts
into independent training replications.

Permutation of task labels holds the learned displacement geometry fixed and
permutes task labels while preserving the number of checkpoints assigned to
each task. Spearman correlations across families are labeled nominal because
task pairs share checkpoints. Exploratory tests across multiple families are
not used as confirmatory evidence.

\subsection{Estimator and Denominator Audit}
\label{sec:e0-audit}

The primary estimator is the ratio of expectations (RoE) for each seed as defined in
the main paper. As a cached sensitivity analysis, we also
compute the expectation of ratios over prompts (EoR),
$\mathbb{E}_{x}[I_x/(M_x+10^{-12})]$, without filtering prompts or changing
the bootstrap draws. The audit reproduces all 68 frozen bootstrap summary
values exactly at tolerance $10^{-12}$. No cell at the family, stratum, and seed level meets
the prospectively specified rule for denominator boundaries. No audited prompt-level
denominator fell below $10^{-4}$, and no denominator-based prompt filtering was applied. Table~\ref{tab:e0-decomposition}
shows that the absolute numerator and denominator vary substantially by
family and stratum, so the main result is not inferred from a contrast alone. Per-seed contrasts under both estimators appear in Table~\ref{tab:e0-estimators}.

Two further controls localize what drives the primary contrast. It is not
vector norm: effects at natural radii and after norm matching are similar on
code and instruction prompts. Nor is it a shrinking denominator: on code
prompts the interaction numerator grows by a factor of $4.7$ while the
base-displacement denominator grows by only $1.4$; on math prompts the two
grow in proportion, which is why the normalized contrast is null there.

\begin{table*}[t]
\centering
\footnotesize
\setlength{\tabcolsep}{4.0pt}
\begin{tabular}{llrrrr}
\toprule
Stratum & Family & $\bar I$ & $\bar M$ & RoE (\%) & EoR (\%) \\
\midrule
Math & code+math & 0.00228 & 0.08114 & 2.88 & 2.75 \\
Math & code+safety & 0.00554 & 0.18635 & 2.98 & 3.30 \\
Code & code+math & 0.00352 & 0.14048 & 2.53 & 2.55 \\
Code & code+safety & 0.01663 & 0.19728 & 8.49 & 8.44 \\
Instruction & code+math & 0.00107 & 0.06252 & 1.74 & 2.02 \\
Instruction & code+safety & 0.01628 & 0.17504 & 9.31 & 10.07 \\
Safety & code+math & 0.00197 & 0.04696 & 4.24 & 4.62 \\
Safety & code+safety & 0.01921 & 0.31038 & 6.20 & 6.31 \\
\bottomrule
\end{tabular}
\caption{Cached estimator decomposition for the Qwen2.5-1.5B surface after norm matching.
$\bar I$ and $\bar M$ are means of the numerator and denominator across three seeds;
RoE and EoR are means of the corresponding estimates for each seed. JSD
quantities are in nats.}
\label{tab:e0-decomposition}
\end{table*}

\begin{table*}[t]
\centering
\footnotesize
\setlength{\tabcolsep}{3.0pt}
\begin{tabular}{llrrrr}
\toprule
Stratum & Estimator & Seed 42 & Seed 123 & Seed 456 & Mean [95\% CI] \\
\midrule
Math & RoE & $-0.81$ & $+0.75$ & $+0.35$ & $+0.10$ $[-0.83,0.91]$ \\
Math & EoR & $-0.05$ & $+0.94$ & $+0.77$ & $+0.55$ $[-0.19,1.13]$ \\
Code & RoE & $+6.09$ & $+7.53$ & $+4.27$ & $+5.96$ $[4.14,7.93]$ \\
Code & EoR & $+5.72$ & $+7.97$ & $+4.01$ & $+5.90$ $[3.94,8.11]$ \\
Instruction & RoE & $+7.64$ & $+9.30$ & $+5.79$ & $+7.58$ $[5.59,9.69]$ \\
Instruction & EoR & $+8.31$ & $+9.51$ & $+6.31$ & $+8.04$ $[6.06,10.04]$ \\
Safety & RoE & $+1.70$ & $+4.17$ & $+0.03$ & $+1.97$ $[-0.03,4.14]$ \\
Safety & EoR & $+1.67$ & $+3.69$ & $-0.30$ & $+1.69$ $[-0.26,3.64]$ \\
\bottomrule
\end{tabular}
\caption{Matched code+safety minus code+math contrasts under RoE (primary)
and EoR (sensitivity), in percentage points. Intervals use the same
hierarchical bootstrap over prompts within seeds and shared prompt indices.
The code and instruction sign pattern is stable across estimators; magnitudes
depend on the estimator.}
\label{tab:e0-estimators}
\end{table*}

The cached surface evaluation predates prompt hash fields recorded during
generation and has no protocol file. The estimator audit records a hash
computed later from the locally archived prompt text but does not claim token
provenance from the time of generation. This limits provenance strength, not
the exact reproduction check against the frozen bootstrap outputs. Results in
formats that machines can read, denominator diagnostics, and the reconstructed
protocol accompany the artifact.

\subsection{Audit in Parameter Space}

\begin{table}[t]
\centering
\footnotesize
\setlength{\tabcolsep}{3pt}
\begin{tabular}{lrr}
\toprule
Metric & 1.5B, six tasks & 0.5B, four tasks \\
\midrule
Checkpoints & 18 & 12 \\
Cosine gap & 0.0184 ($p=.0002$) & 0.0045 ($p=.1146$) \\
Silhouette & 0.0122 ($p=.0002$) & $-0.0008$ ($p=.0886$) \\
1-NN accuracy & 0.444 ($p=.0062$) & 0.250 ($p=.3755$) \\
Participation ratio & 16.68 & 10.82 \\
Stable rank & 13.80 & 9.22 \\
\bottomrule
\end{tabular}
\caption{Diagnostics of parameter directions. $p$ values are results from permutation of task labels. Participation ratio and stable rank are descriptive because their attainable values depend on checkpoint count. For Llama-3.1-8B (three tasks), the within-versus-between task cosine gap computed from adapter factors is $0.0385$ ($p=.0032$).}
\label{tab:param-audit}
\end{table}

\subsection{Prospectively Specified Unseen Families}

The complete ordering of nine families appears in the main paper's OOS table;
the matrix over pairs, strata, and seeds accompanies the artifact in a format that machines can read.
The primary comparison counted all registered pairs from the high and low bins, producing
8/8 preregistered signs. Adding the prospectively specified code+math low anchor gives 12/12.
The bridge exceeds both low families (2/2), whereas high exceeds middle
in 5/8 comparisons. Accuracy for high and low comparisons within each prompt stratum is 37.5\%
(math), 50\% (code), 100\% (instruction), 100\% (safety), 100\%
(summarization), and 75\% (translation).

\paragraph{Exact design-preserving randomization test.}
The only randomization that preserves this design relabels the four
calibration-era tasks, whose prospectively specified bins form the multiset
$\{\text{High},\text{High},\text{Middle},\text{Low}\}$. Because two labels are
identical there are $4!/2!=12$ distinct assignments, each yielding the same
$4\times2=8$ high-versus-low family comparisons. Enumerating all 12, the
observed assignment is the unique maximum at 8/8; the next highest assignments
reach 7/8. The exact one-sided $p$-value is therefore $1/12=0.083$, which is
also the smallest value this design can attain. The test reports the
finite-sample resolution of the prospective specification rather than a
definitive inferential result. The enumeration script and its
machine-readable outputs, listing every assignment with its hit count, are
included with the artifact.

\subsection{Matched Contrast Across Scales}

Table~\ref{tab:cross-scale-full} reports the full four-stratum matched contrasts at 1.5B and 0.5B.

\begin{table}[t]
\centering
\footnotesize
\begin{tabular}{lrr}
\toprule
Stratum & 1.5B contrast & 0.5B contrast \\
\midrule
Math & $+0.10$ $[-0.83,0.91]$ & $+0.50$ $[-0.08,1.31]$ \\
Code & $+5.96$ $[4.14,7.93]$ & $+6.21$ $[3.11,9.04]$ \\
Instruction & $+7.58$ $[5.59,9.69]$ & $+5.26$ $[3.74,6.70]$ \\
Safety & $+1.97$ $[-0.03,4.14]$ & $+5.52$ $[3.21,8.26]$ \\
\bottomrule
\end{tabular}
\caption{Interaction ratio contrasts between code+safety and code+math after norm matching, in percentage points with 95\% intervals from the hierarchical bootstrap.}
\label{tab:cross-scale-full}
\end{table}

At 0.5B, all three seed differences are positive for code
($+7.45,+3.27,+7.91$ points) and instruction
($+6.17,+3.70,+5.90$ points).  This test was frozen before inspecting the 0.5B
response-only surfaces.

\subsection{Public Prompts, Format, and Execution Boundary}
\label{sec:public-boundary-appendix}

The public prompt analysis evaluates raw prompt sets for code, math,
instruction, and safety. The public math contrast after norm matching is $+0.36$
points and satisfies
the practical equivalence gate. Raw public code, instruction, and safety
prompts retain positive contrasts of $+6.91$, $+11.09$, and
$+3.96$ points after norm matching, respectively. A format bridge diagnostic then wraps the
identical 60 public code prompts in the Alpaca instruction template; the
contrast after norm matching collapses to $+0.34$ points. The public boundary
depends on prompt format rather than reflecting a simple failure on public distributions.

The secondary measurement over sequences was frozen before any result on public prompts
was observed and uses reference continuations through teacher forcing where such
references are available. On public code references, the contrast over sequences
after norm matching is $+3.72$ points with interval $[2.77,4.85]$, reproducing
the direction at the first token with lower magnitude. On public math references, the
contrast over sequences after norm matching is $+1.87$ points with interval
$[0.88,2.93]$. The equivalence conclusion for public math is specific to the
primary estimand at the first token; at sequence resolution, a small but detectable
residual remains.

For execution evaluation, let $S_0,S_a,S_b,S_{ab}$ be pass@1 for the base,
endpoints for each task, and merged model. The signed factorial interaction is
$I_S=S_{ab}-S_a-S_b+S_0$.  We compare $|I_S|$ between code+safety and code+math.
Across three HumanEval+ seeds, the absolute matched contrast is $+4.14$ points
with interval $[-2.44,12.60]$; MBPP+ is approximately zero.  These outcomes do
not support a claim about predicting benchmark performance.

\subsection{Compute and Artifact Plan}

The 1.5B replication and expansion were run in a Linux container on an NVIDIA
GeForce RTX 5090 with 32~GB memory; environment validation logs record the
exact PyTorch/CUDA versions and bfloat16 support.  The initial 0.5B pilot used
an NVIDIA A100 with 40~GB memory.  Exact environment lock files,
configurations, prompt strata, prospective prediction documents, figure source
tables, analysis scripts, and checkpoint identifiers will accompany the public
artifact. Because Llama-3.1 weights are governed by the Llama 3.1 Community
License, the artifact will not redistribute the base checkpoint; it will provide
the locked model revision, adapter metadata, and analysis outputs.  The released
artifact will distinguish the earlier pilot that masked EOS from the final
response-only protocol.

\section{Claim Boundary Checklist}

The evidence supports functional interaction that depends on the input, coarse
ordering of unseen families, persistence under fine-tuning of all parameters
at 0.5B, persistence across Qwen2.5 scales, and a directed audit on
Llama-3.1-8B as a second architecture. It does not support a global semantic
coordinate system, universal architecture generality, broad generality under
fine-tuning of all parameters, prediction of interference for individual
examples, or prediction of downstream performance. Adapters trained for
safety are used as task families; the study does not measure harmful
compliance or excessive refusal and makes no claim of guaranteed safety.